# Deep Image Matting

Ning Xu[1,2], Brian Price[3], Scott Cohen[3], and Thomas Huang[1,2]

[1]Beckman Institute for Advanced Science and Technology
[2]University of Illinois at Urbana-Champaign
[3]Adobe Research
{*ningxu2,t-huang1*}@*illinois.edu*, {*bprice,scohen*}@*adobe.com*

**Abstract**

*Image matting is a fundamental computer vision problem and has many applications. Previous algorithms have poor performance when an image has similar foreground and background colors or complicated textures. The main reasons are prior methods 1) only use low-level features and 2) lack high-level context. In this paper, we propose a novel deep learning based algorithm that can tackle both these problems. Our deep model has two parts. The first part is a deep convolutional encoder-decoder network that takes an image and the corresponding trimap as inputs and predict the alpha matte of the image. The second part is a small convolutional network that refines the alpha matte predictions of the first network to have more accurate alpha values and sharper edges. In addition, we also create a large-scale image matting dataset including 49300 training images and 1000 testing images. We evaluate our algorithm on the image matting benchmark, our testing set, and a wide variety of real images. Experimental results clearly demonstrate the superiority of our algorithm over previous methods.*

## 1. Introduction

Matting, the problem of accurate foreground estimation in images and videos, has significant practical importance. It is a key technology in image editing and film production and effective natural image matting methods can greatly improve current professional workflows. It necessitates methods that handle real world images in unconstrained scenes.

Unfortunately, current matting approaches do not generalize well to typical everyday scenes. This is partially due to the difficulty of the problem: as formulated the matting problem is underconstrained with 7 unknown values per pixel but only 3 known values:

$$I_i = \alpha_i F_i + (1-\alpha_i) B_i \quad \alpha_i \in [0,1]. \tag{1}$$

where the RGB color at pixel $i$, $I_i$, is known and the foreground color $F_i$, background color $B_i$ and matte estimation $\alpha_i$ are unknown. However, current approaches are further limited in their approach.

The first limitation is due to current methods being designed to solve the matting equation (Eq. 1). This equation formulates the matting problem as a linear combination of two colors, and consequently most current algorithms approach this largely as a color problem. The standard approaches include sampling foreground and background colors [3, 9], propagating the alpha values according to the matting equation [14, 31, 22], or a hybrid of the two [32, 13, 28, 16]. Such approaches rely largely on color as the distinguishing feature (often along with the spatial position of the pixels), making them incredibly sensitive to situations where the foreground and background color distributions overlap, which unfortunately for these methods is the common case for natural images, often leading to low-frequency "smearing" or high-frequency "chunky" artifacts depending on the method (see Fig 1 top row). Even the recently proposed deep learning methods are highly reliant on color-dependent propagation methods [8, 29].

A second limitation is due to the focus on a very small dataset. Generating ground truth for matting is very difficult, and the alphamatting.com dataset [25] made a significant contribution to matting research by providing ground-truth data. Unfortunately, it contains only 27 training images and 8 test images, most of which are objects in front of an image on a monitor. Due to its size and constraints of the dataset (e.g. indoor lab scenes, indoor lighting, no humans or animals), it is by its nature biased, and methods are incentivized to fit to this data for publication purposes. As is the case with all datasets, especially small ones, at some point methods will overfit to the dataset and no longer generalize to real scenes. A recent video matting dataset is available [10] with 3 training videos and 10 test videos, 5 of which were extracted from green screen footage and the

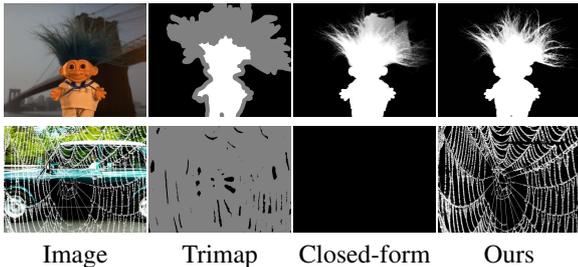

Image    Trimap    Closed-form    Ours

Figure 1. The comparison between our method and the Closed-form matting [22]. The first image is from the Alpha Matting benchmark and the second image is from our 1000 testing images.

rest using a similar method to [25].

In this work, we present an approach aimed to overcome these limitations. Our method uses deep learning to directly compute the alpha matte given an input image and trimap. Instead of relying primarily on color information, our network can learn the natural structure that is present in alpha mattes. For example, hair and fur (which usually require matting) possess strong structural and textural patterns. Other cases requiring matting (e.g. edges of objects, regions of optical or motion blur, or semi-transparent regions) almost always have a common structure or alpha profile that can be expected. While low-level features will not capture this structure, deep networks are ideal for representing it. Our two-stage network includes an encoder-decoder stage followed by a small residual network for refinement and includes a novel composition loss in addition to a loss on the alpha. We are the first to demonstrate the ability to learn an alpha matte end-to-end given an image and trimap.

To train a model that will excel in natural images of unconstrained scenes, we need a much larger dataset than currently available. Obtaining a ground truth dataset using the method of [25] would be very costly and cannot handle scenes with any degree of motion (and consequently cannot capture humans or animals). Instead, inspired by other synthetic datasets that have proven sufficient to train models for use in real images (e.g. [4]), we create a large-scale matting dataset using composition. Images with objects on simple backgrounds were carefully extracted and were composited onto new background images to create a dataset with 49300 training images and 1000 test images.

We perform extensive evaluation to prove the effectiveness on our method. Not only does our method achieve first place on the alphamatting.com challenge, but we also greatly outperform prior methods on our synthetic test set. We show our learned model generalizes to natural images with a user study comparing many prior methods on 31 natural images featuring humans, animals, and other objects in varying scenes and under different lighting conditions. This study shows a strong preference for our results, but also shows that some methods which perform well on the alphamatting.com dataset actually perform worse compared to other methods when judged by humans, suggesting that methods are being to overfit on the alphamatting.com test set. Finally, we also show that we are more robust to trimap placement than other methods. In fact, we can produce great results even when there is no known foreground and/or background in the trimap while most methods cannot return any result (see Fig 1 bottom row ).

## 2. Related works

Current matting methods rely primarily on color to determine the alpha matte, along with positional or other low-level features. They do so through sampling, propagation, or a combination of the two.

In sampling-based methods [3, 9, 32, 13, 28, 16], the known foreground and background regions are sampled to find candidate colors for a given pixel's foreground and background, then a metric is used to determine the best foreground/background combination. Different sampling methods are used, including sampling along the boundary nearest the given pixel [32], sampling based on ray casting [13], searching the entire boundary [16], or sampling from color clusters [28, 12]. The metric to decide among the sampled candidate nearly always includes a matting equation reconstruction error, potentially with terms measuring the distance of samples from the given pixel [32, 16] or the similarity of the foreground and background samples [32, 28], and formulations include sparse coding [12] and KL-divergence approaches [19, 18]. Higher-order features like texture [27] have been used rarely and have limited effectiveness.

In propagation methods, Eq. 1 is reformulated such that it allows propagation of the alpha values from the known foreground and background regions into the unknown region. A popular approach is Closed-form Matting [22] which is often used as a post-process after sampling [32, 16, 28]. It derives a cost function from local smoothness assumption on foreground and background colors and finds the globally optimal alpha matte by solving a sparse linear system of equations. Other propagation methods include random walks [14], solving Poisson equations [31], and nonlocal propagation methods [21, 7, 5].

Recently, several deep learning works have been proposed for image matting. However, they do not directly learn an alpha matte given an image and trimap. Shen et al. [29] use deep learning for creating a trimap of a person in a portrait image and use [22] for matting through which matting errors are backpropagated to the network. Cho et al. [8] take the matting results of [22] and [5] and normalized RGB colors as inputs and learn an end-to-end deep network to predict a new alpha matte. Although both our algorithm and the two works leverage deep learning, our algorithm is quite different from theirs. Our algorithm directly learns the alpha matte given an image and trimap while the other two works rely on existing algorithms to compute the actual matting, making their methods vulnerable to the same problems as previous matting methods.

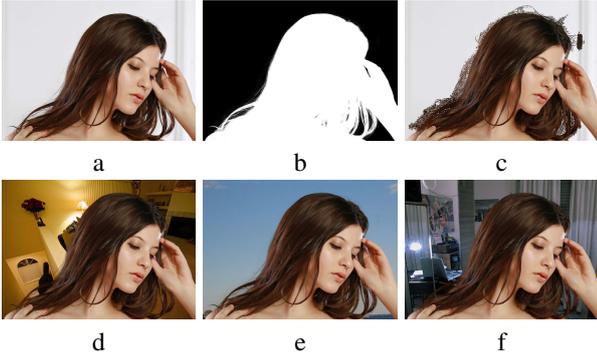

Figure 2. Dataset creation. (a) An input image with a simple background is matted manually. The (b) computed alpha matte and (c) computed foreground colors are used as ground truth to composite the object onto (d-f) various background images.

## 3. New matting dataset

The matting benchmark on alphamatting.com [25] has been tremendously successful in accelerating the pace of research in matting. However, due to the carefully controlled setting required to obtain ground truth images, the dataset consists of only 27 training images and 8 testing images. Not only is this not enough images to train a neural network, but it is severely limited in its diversity, restricted to small-scale lab scenes with static objects.

To train our matting network, we create a larger dataset by compositing objects from real images onto new backgrounds. We find images on simple or plain backgrounds (Fig. 2a), including the 27 training images from [25] and every fifth frame from the videos from [26]. Using Photoshop, we carefully manually create an alpha matte (Fig. 2b) and pure foreground colors (Fig. 2c). Because these objects have simple backgrounds we can pull accurate mattes for them. We then treat these as ground truth and for each alpha matte and foreground image, we randomly sample $N$ background images in MS COCO [23] and Pascal VOC [11], and composite the object onto those background images.

We create both a training and a testing dataset in the above way. Our training dataset has 493 unique foreground objects and 49,300 images ($N = 100$) while our testing dataset has 50 unique objects and 1000 images ($N = 20$). The trimap for each image is randomly dilated from its ground truth alpha matte. In comparison to previous matting datasets, our new dataset has several advantages. 1) It has many more unique objects and covers various matting cases such as hair, fur, semi-transparency, etc. 2) Many composited images have similar foreground and background colors and complex background textures, making our dataset more challenging and practical.

An early concern is whether this process would create a bias due to the composited nature of the images, such that a network would learn to key on differences in the foreground and background lighting, noise levels, etc. However, we found experimentally that we achieved far superior results on natural images compared to prior methods (see Sec. 5.3).

## 4. Our method

We address the image matting problem using deep learning. Given our new dataset, we train a neural network to fully utilize the data. The network consists of two stages (Fig. 3). The first stage is a deep convolutional encoder-decoder network which takes an image patch and a trimap as input and is penalized by the alpha prediction loss and a novel compositional loss. The second stage is a small fully convolutional network which refines the alpha prediction from the first network with more accurate alpha values and sharper edges. We will describe our algorithm with more details in the following sections.

### 4.1. Matting encoder-decoder stage

The first stage of our network is a deep encoder-decoder network (see Fig. 3), which has achieved successes in many other computer vision tasks such as image segmentation [2], boundary prediction [33] and hole filling [24].

**Network structure:** The input to the network is an image patch and the corresponding trimap which are concatenated along the channel dimension, resulting in a 4-channel input. The whole network consists of an encoder network and a decoder network. The input to the encoder network is transformed into downsampled feature maps by subsequent convolutional layers and max pooling layers. The decoder network in turn uses subsequent unpooling layers which reverse the max pooling operation and convolutional layers to upsample the feature maps and have the desired output, the alpha matte in our case. Specifically, our encoder network has 14 convolutional layers and 5 max-pooling layers. For the decoder network, we use a smaller structure than the encoder network to reduce the number of parameters and speed up the training process. Specifically, our decoder network has 6 convolutional layers, 5 unpooling layers followed by a final alpha prediction layer.

**Losses:** Our network leverages two losses. The first loss is called the alpha-prediction loss, which is the absolute difference between the ground truth alpha values and the predicted alpha values at each pixel. However, due to the non-differentiable property of absolute values, we use the following loss function to approximate it.

$$\mathcal{L}_\alpha^i = \sqrt{(\alpha_p^i - \alpha_g^i)^2 + \epsilon^2}, \quad \alpha_p^i, \alpha_g^i \in [0,1]. \quad (2)$$

where $\alpha_p^i$ is the output of the prediction layer at pixel $i$

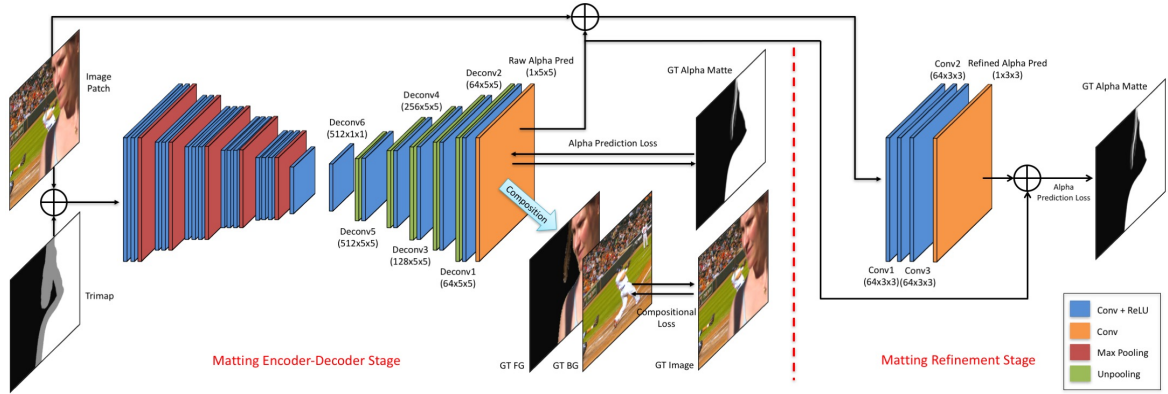

Figure 3. Our network consists of two stages, an encoder-decoder stage (Sec. 4.1) and a refinement stage (Sec. 4.2)

thresholded between 0 and 1. $\alpha_g^i$ is the ground truth alpha value at pixel $i$. $\epsilon$ is a small value which is equal to $10^{-6}$ in our experiments. The derivative $\frac{\partial \mathcal{L}_\alpha^i}{\partial \alpha_p^i}$ is straightforward.

$$\frac{\partial \mathcal{L}_\alpha^i}{\partial \alpha_p^i} = \frac{\alpha_p^i - \alpha_g^i}{\sqrt{(\alpha_p^i - \alpha_g^i)^2 + \epsilon^2}}. \quad (3)$$

The second loss is called the compositional loss, which is the absolute difference between the ground truth RGB colors and the predicted RGB colors composited by the ground truth foreground, the ground truth background and the predicted alpha mattes. Similarly, we approximate it by using the following loss function.

$$\mathcal{L}_c^i = \sqrt{(c_p^i - c_g^i)^2 + \epsilon^2}. \quad (4)$$

where $c$ denotes the RGB channel, $p$ denotes the image composited by the predicted alpha, and $g$ denotes the image composited by the ground truth alphas. The compositional loss constrains the network to follow the compositional operation, leading to more accurate alpha predictions.

The overall loss is the weighted summation of the two individual losses, i.e., $\mathcal{L}_{overall} = w_l \cdot \mathcal{L}_\alpha + (1 - w_l) \cdot \mathcal{L}_c$, where $w_l$ is set to 0.5 in our experiment. In addition, since only the alpha values inside the unknown regions of trimaps need to be inferred, we therefore set additional weights on the two types of losses according to the pixel locations, which can help our network pay more attention on the important areas. Specifically, $w_i = 1$ if pixel $i$ is inside the unknown region of the trimap while $w_i = 0$ otherwise.

**Implementation:** Although our training dataset has 49,300 images, there are only 493 unique objects. To avoid overfitting as well as to leverage the training data more effectively, we use several training strategies. First, we randomly crop 320×320 (image, trimap) pairs centered on pixels in the unknown regions. This increases our sampling space. Second, we also crop training pairs with different sizes (e.g. 480×480, 640×640) and resize them to 320×320. This makes our method more robust to scales and helps the network better learn context and semantics. Third, flipping is performed randomly on each training pair. Fourth, the trimaps are randomly dilated from their ground truth alpha mattes, helping our model to be more robust to the trimap placement. Finally, the training inputs are recreated randomly after each training epoch.

The encoder portion of the network is initialized with the first 14 convolutional layers of VGG-16 [30] (the 14th layer is the fully connected layer "fc6" which can be transformed to a convolutional layer). Since the network has 4-channel input, we initialize the one extra channel of the first-layer convolutional filters with zeros. All the decoder parameters are initialized with Xavier random variables.

When testing, the image and corresponding trimap are concatenated as the input. A forward pass of the network is performed to output the alpha matte prediction. When a GPU memory is insufficient for large images, CPU testing can be performed.

### 4.2. Matting refinement stage

Although the alpha predictions from the first part of our network are already much better than existing matting algorithms, because of the encoder-decoder structure, the results are sometimes overly smooth. Therefore, we extend our network to further refine the results from the first part. This extended network usually predicts more accurate alpha mattes and sharper edges.

**Network structure:** The input to the second stage of our network is the concatenation of an image patch and its alpha prediction from the first stage (scaled between 0 and 255), resulting in a 4-channel input. The output is the corresponding ground truth alpha matte. The network is a fully convolutional network which includes 4 convolutional layers. Each of the first 3 convolutional layers is followed by a non-linear "ReLU" layer. There are no downsampling layers since we want to keep very subtle structures missed in the first stage. In addition, we use a "skip-model" structure where the 4-th channel of the input data is first scaled between 0 and 1 and then is added to the output of the network. The detailed configuration is shown in Fig. 3.

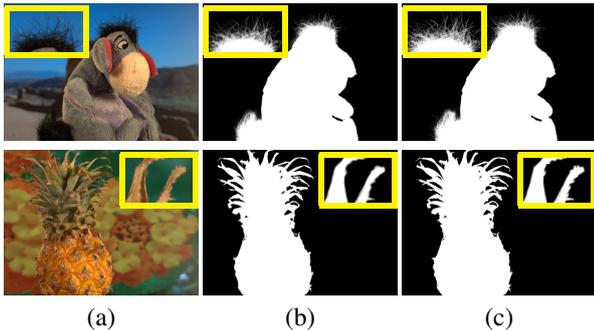

Figure 4. The effect of our matting refinement network. (a) The input images. (b) The results of our matting encoder-decoder stage. (c) The results of our matting refinement stage.

Table 1. The quantitative results on the Composition-1k testing dataset. The variants of our approaches are emphasized in italic. The best results are emphasized in bold.

| Methods | SAD | MSE | Gradient | Connectivity |
|---|---|---|---|---|
| Shared Matting [13] | 128.9 | 0.091 | 126.5 | 135.3 |
| Learning Based Matting [34] | 113.9 | 0.048 | 91.6 | 122.2 |
| Comprehensive Sampling [28] | 143.8 | 0.071 | 102.2 | 142.7 |
| Global Matting [16] | 133.6 | 0.068 | 97.6 | 133.3 |
| Closed-Form Matting [22] | 168.1 | 0.091 | 126.9 | 167.9 |
| KNN Matting [5] | 175.4 | 0.103 | 124.1 | 176.4 |
| DCNN Matting [8] | 161.4 | 0.087 | 115.1 | 161.9 |
| *Encoder-Decoder network (single alpha prediction loss)* | 59.6 | 0.019 | 40.5 | 59.3 |
| *Encoder-Decoder network* | 54.6 | 0.017 | 36.7 | 55.3 |
| *Encoder-Decoder network + Guided filter[17]* | 52.2 | 0.016 | **30.0** | 52.6 |
| *Encoder-Decoder network + Refinement network* | **50.4** | **0.014** | 31.0 | **50.8** |

The effect of our refinement stage is illustrated in Fig. 4. Note that it does not make large-scale changes to the alpha matte, but rather just refines and sharpens the alpha values.

**Implementation:** During training, we first update the encoder-decoder part without the refinement part. After the encoder-decoder part is converged, we fix its parameters and then update the refinement part. Only the alpha prediction loss (Eqn. 2) is used due to its simple structure. We also use all the training strategies of the 1st stage except the 4th one. After the refinement part is also converged, finally we fine-tune the the whole network together. We use Adam [20] to update both parts. A small learning rate $10^{-5}$ is set constantly during the training process.

During testing, given an image and a trimap, our algorithm first uses the matting encoder-decoder stage to get an initial alpha matte prediction. Then the image and the alpha prediction are concatenated as the input to the refinement stage to produce the final alpha matte prediction.

## 5. Experimental results

In this section we evaluate our method on 3 datasets. 1) We evaluate on the alphamatting.com dataset [25], which is the existing benchmark for image matting methods. It includes 8 testing images, each has 3 different trimaps, namely, "small", "large" and "user". 2) Due to the limited size and range of objects in the alphamatting.com dataset, we propose the Composition-1k test set. Our composition-based dataset includes 1000 images and 50 unique foregrounds. This dataset has a wider range of object types and background scenes. 3) To measure our performance on natural images, we also collect a third dataset including 31 natural images. The natural images cover a wide range of common matting foregrounds such as person, animals, *etc*.

### 5.1. The alphamatting.com dataset

Our approach achieves the top results compared to all the other methods on the alphamatting.com benchmark. Specifically, our method ranks the 1st place in terms of the SAD metric. Our method also has the smallest SAD errors for 5 images with all the 3 trimaps (Fig. 5). In addition, our method ranks the 2nd place in terms of both the MSE and Gradient metrics. Overall, our method is one of the best performers on this dataset.

A key reason for our success is our network's ability to learn structure and semantics, which is important for the accurate estimation of alpha matte when the background scene is complex or the background and foreground colors are similar. For example, in Fig 6 the "Troll" example has very similar colors of the hair and the bridge while the "Doll" example has strong textured background. The best results of previous methods (from column 3 to column 6) all have very obvious mistakes in those hard regions. In contrast, our method directly learns object structure and image context. As a result, our method not only avoids the similar mistakes made by previous methods but also predicts more details. It is worth noting that although DCNN matting [8] is also a deep-learning based method, it learns the non-linear combination of previous matting methods within small local patches. Therefore the method cannot really understand semantics and thus has the same limitations as previous non-deep-learning-based methods.

### 5.2. The Composition-1k testing dataset

We further evaluate 7 top performing prior methods and each component of our approach on the Composition-1k testing dataset. For all prior methods, the authors' provided codes are used. The different variants of our approach include: the matting encoder-decoder network 1) with only the alpha prediction loss, 2) with both the alpha prediction loss and the compositional loss, the matting encoder-decoder network 3) post-processed by the Guided filter [17] and 4) post-processed by the matting refinement network.

| Sum of Absolute Differences | overall rank | avg. small rank | avg. large rank | avg. user rank | Troll (Strongly Transparent) Input | | | Doll (Strongly Transparent) Input | | | Donkey (Medium Transparent) Input | | | Elephant (Medium Transparent) Input | | | Plant (Little Transparent) Input | | | Pineapple (Little Transparent) Input | | | Plastic bag (Highly Transparent) Input | | | Net (Highly Transparent) Input | | |
|---|---|---|---|---|---|---|---|---|---|---|---|---|---|---|---|---|---|---|---|---|---|---|---|---|---|---|---|---|
| | | | | | small | large | user | small | large | user | small | large | user | small | large | user | small | large | user | small | large | user | small | large | user | small | large | user |
| Deep Matting | 2.4 | 3.1 | 1.8 | 2.4 | 10.7 1 | 11.2 1 | 11 1 | 4.8 1 | 5.8 1 | 5.6 1 | 2.8 1 | 2.9 1 | 2.9 1 | 1.1 1 | 1.1 1 | 2 1 | 6 11 | 7.1 2 | 8.9 1 | 2.7 1 | 3.2 1 | 3.9 1 | 19.2 2 | 19.6 3 | 18.7 4 | 21.8 7 | 23.9 4 | 24.1 9 |
| DCNN Matting | 3.3 | 4.6 | 1.8 | 3.5 | 12 9 | 14.1 3 | 14.5 5 | 5.3 2 | 6.4 2 | 6.8 4 | 3.9 2 | 4.5 2 | 3.4 3 | 1.6 7 | 2.5 2 | 2.2 4 | 6 10 | 6.9 1 | 9.1 2 | 4 3 | 6 2 | 5.3 3 | 19.9 3 | 19.2 1 | 19.1 6 | 19.4 1 | 20 1 | 21.2 1 |
| CSC Matting | 10 | 13.5 | 6.4 | 10.3 | 13.6 22 | 15.6 4 | 14.5 4 | 6.2 12 | 7.5 5 | 8.1 15 | 4.6 14 | 4.8 4 | 4.2 18 | 1.8 12 | 2.7 3 | 2.5 8 | 5.5 4 | 7.3 3 | 9.7 3 | 4.6 8 | 7.6 5 | 6.9 12 | 23.7 12 | 23 11 | 21 11 | 26.3 24 | 27.2 16 | 25.2 11 |
| LNSP Matting | 10.7 | 7.4 | 10.3 | 14.4 | 12.2 10 | 22.5 27 | 19.5 32 | 5.6 3 | 8.1 7 | 8.8 24 | 4.6 12 | 5.9 19 | 3.6 5 | 1.5 4 | 3.5 10 | 3.1 19 | 6.2 12 | 8.1 7 | 10.7 7 | 4 2 | 7.1 3 | 6.4 8 | 21.5 6 | 20.8 4 | 16.3 1 | 22.5 10 | 24.4 5 | 27.8 19 |
| Graph-based sparse matting | 11.1 | 11.5 | 11.5 | 10.3 | 12.6 15 | 20.5 19 | 14.8 9 | 5.7 6 | 7.3 3 | 6.4 3 | 4.5 10 | 5.3 11 | 3.7 6 | 1.4 3 | 3.3 8 | 2.3 6 | 6.3 14 | 7.9 4 | 11.1 8 | 4.2 4 | 8.3 9 | 6.4 7 | 28.7 28 | 31.3 29 | 27.1 25 | 23.6 12 | 25.1 9 | 27.3 18 |

Figure 5. SAD results on the alphamatting.com dataset. The top 5 methods are shown. Our method is emphasized by a red rectangle.

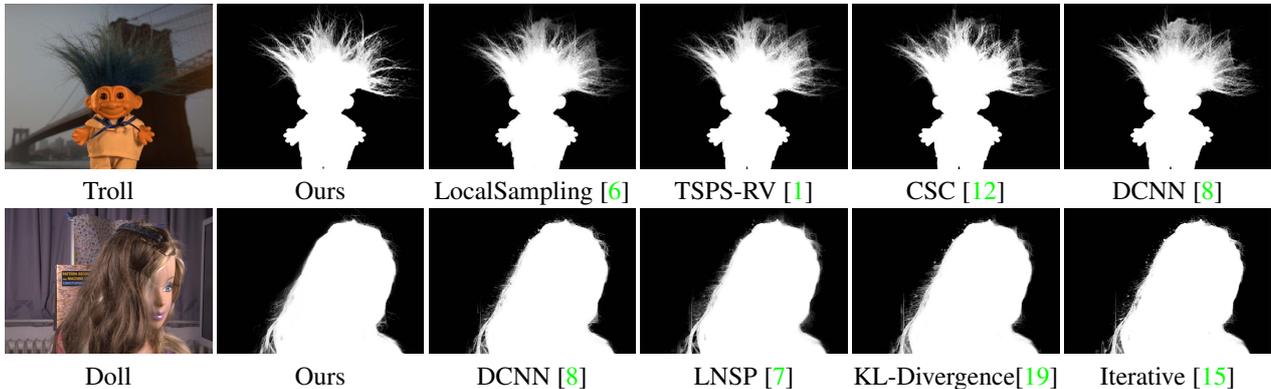

| Troll | Ours | LocalSampling [6] | TSPS-RV [1] | CSC [12] | DCNN [8] |

| Doll | Ours | DCNN [8] | LNSP [7] | KL-Divergence[19] | Iterative [15] |

Figure 6. The alpha matte predictions of the test images "Troll" with trimap "user" and "Doll" with trimap "small". The first column shows the test images. For each test image, the 1st ranked result to the 5th ranked result under the SAD metric are displayed from column two to column six in decreasing orders. In both examples, our method achieves the best results.

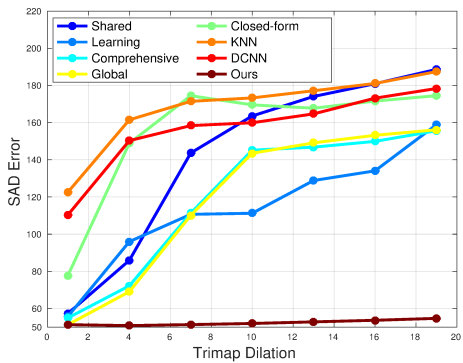

Figure 7. The SAD error at different levels of trimap dilation.

The quantitative results under the SAD, MSE, Gradient and Connectivity errors proposed by [25] are displayed in Table 1. Clearly all variants of our approach have much better results than the other methods. The main reason is still the capability of our deep model understanding the complex context of images while the other methods cannot. By comparing the variants of our approach, we can also validate the effectiveness of each component of our approach: 1) the compositional loss helps our model learn the compositional operation, and thus leads to better results, 2) the results of our matting encoder-decoder network can be improved by combining with previous edge-preserving filters (*e.g.* Guided filter [17]) as well as our matting refinement network. But the latter one has more obvious improvement both visually and quantitatively since it is directly trained with the outputs of our encoder-decoder network.

We test the sensitivity of our method to trimap placement in Fig. 7. We evaluate over a subset of our dataset that includes one randomly-chosen image for each unique object for a total of 50 images. To form the trimap, we dilate the ground truth alpha for each image by $d$ pixels for increasing values of $d$. The SAD errors at a particular parameter $d$ are averaged over all images. The results of all the methods at parameters $d \in [1, 4, 7, 10, 13, 16, 19]$ are shown in Fig. 7. Clearly our method has a low and stable error rate with the increasing values of $d$ whiles the error rate of the other approaches increases rapidly. Our good performance derives from both our training strategies as well as a good understanding of image context.

Some visual examples are shown in Fig. 8 to demonstrate the good performance of our approach on different matting cases such as hair, holes and semi-transparency. Moreover, our approach can also handle objects with no pure foreground pixels, as shown in the last example in Fig. 8. Since previous sampling-based and propagation-based methods must leverage known foreground and background pixels, they cannot handle this case, while our approach can learn the appearance of fine details directly from data.

### 5.3. The real image dataset

Matting methods should generalize well to real-world images. To validate the performance of our approach and other methods on real images, we conduct a user study on the real image dataset. These images consist of images pulled from the internet as well as images provided by the ICCV 2013 tutorial on image matting.

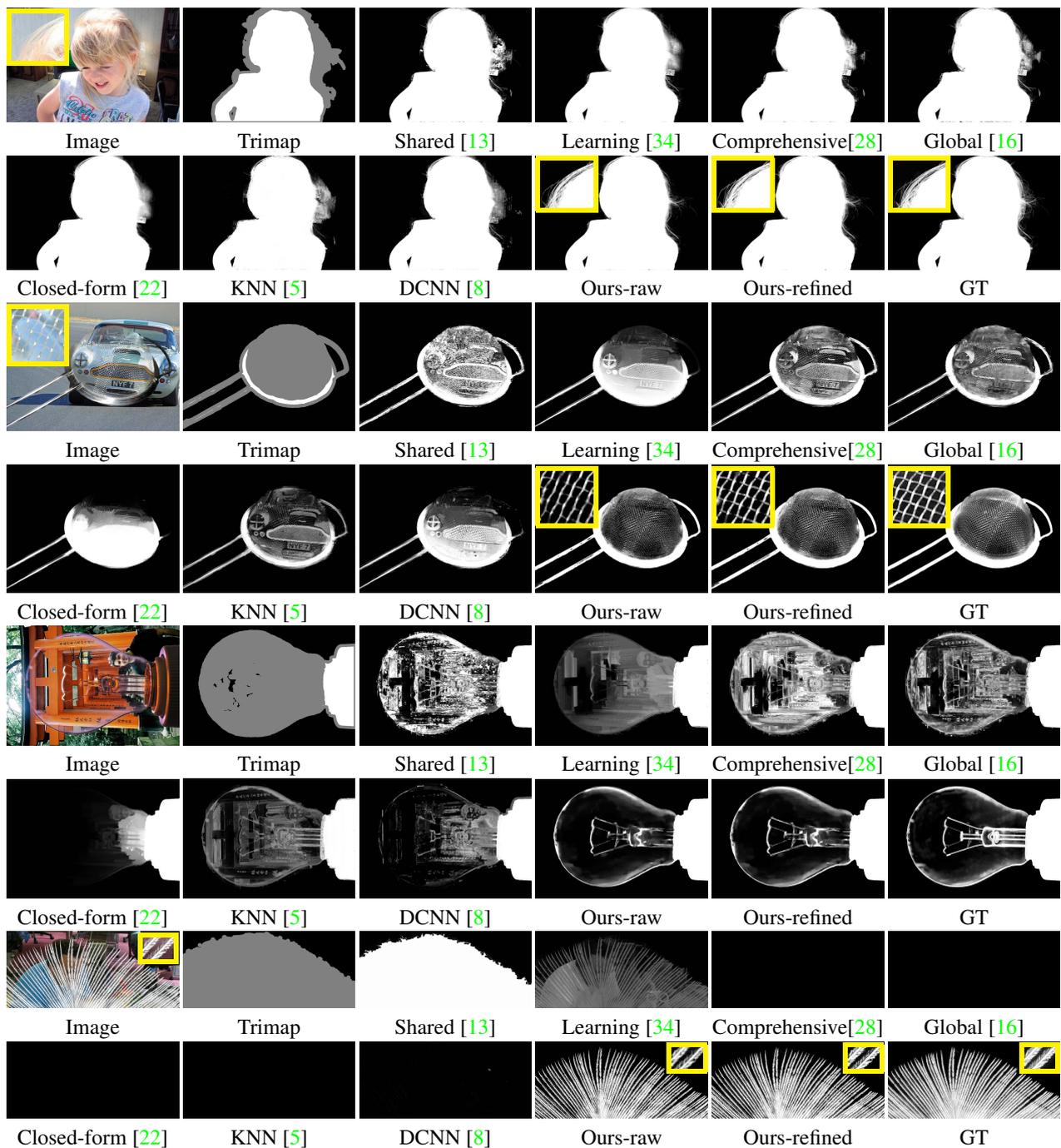

Figure 8. The visual comparison results on the Composition-1k testing dataset. "Ours-raw" denotes the results of our matting encoder-decoder stage while "Ours-refined" denotes the results of our matting refinement stage.

Because our subjects may not be acquainted with alpha mattes, we instead evaluate the results of compositions. For each method, the computed alpha matte is used to blend the test image onto a black background and onto a white background. For the user test, we present the image and the two composition results of two randomly selected approaches to an user and ask which results are more accurate and realistic especially in the regions of fine details (*e.g.* hair, edges of object, and semi-transparent areas). To avoid evaluation bias, we conduct the user study on the Amazon Mechanical Turk. As a result, there are total 392 users participating the user study and each method pair on one image is evaluated by 5 to 6 unique users.

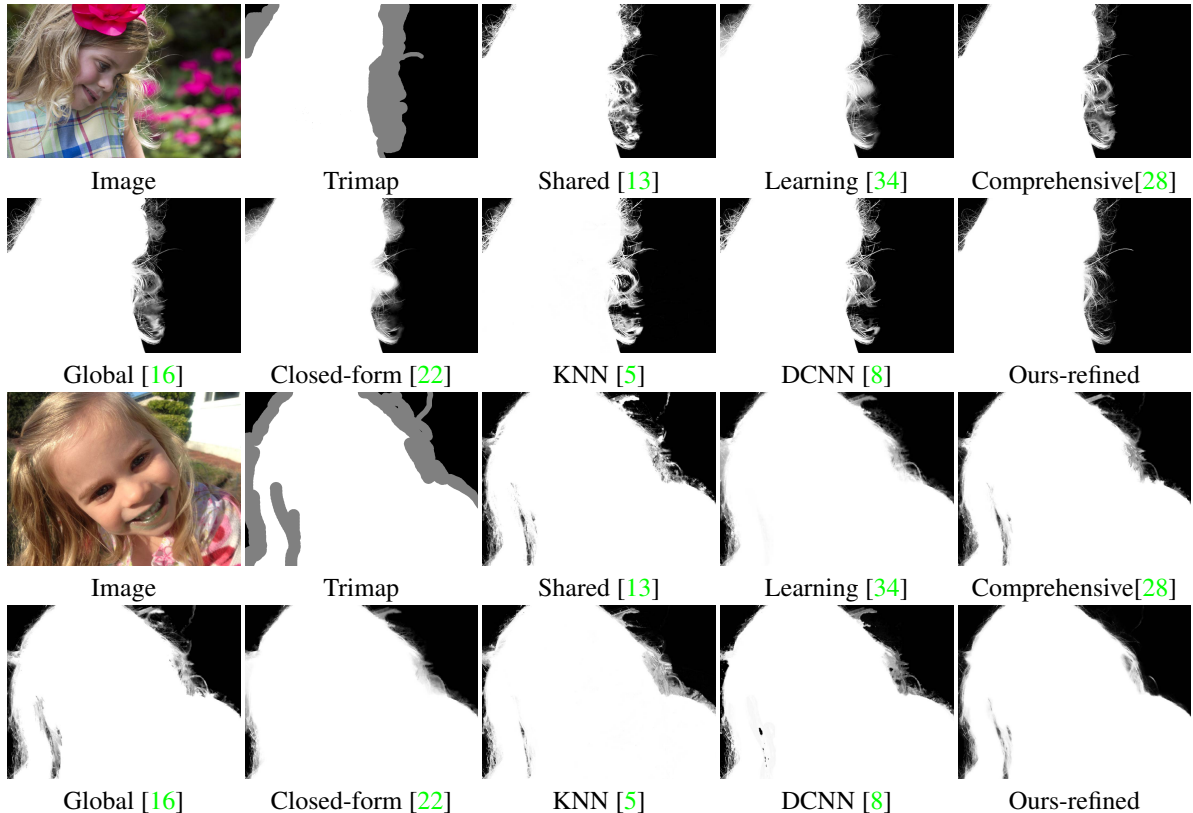

Figure 9. Example results from our real image dataset.

Table 2. The user study on the real image dataset. The preferred method in each pairwise comparison is emphasized in bold.

| Methods | [13] | [34] | [28] | [16] | [22] | [5] | [8] | Ours |
|---|---|---|---|---|---|---|---|---|
| Shared [13] | - | **60.0** | **78.5** | **79.6** | **69.7** | 40.6 | **57.8** | **83.7** |
| Learning [34] | 40.0 | - | **60.2** | **54.6** | **53.4** | 27.3 | 35.1 | **83.6** |
| Comprehensive [28] | 21.5 | 39.8 | - | 25.8 | 43.3 | 20.4 | 29.2 | **78.8** |
| Global [16] | 20.4 | 45.4 | **74.2** | - | **53.3** | 30.0 | 42.0 | **84.2** |
| Closed-Form [22] | 30.3 | 46.6 | **56.7** | 46.7 | - | 25.0 | 38.1 | **80.4** |
| KNN [5] | **59.4** | **72.7** | **79.6** | **70.0** | **75.0** | - | **73.3** | **97.0** |
| DCNN [8] | 42.2 | **64.9** | **70.8** | **58.0** | **61.9** | 26.7 | - | **83.7** |
| Ours | 16.3 | 16.4 | 21.2 | 15.8 | 19.6 | 3.0 | 16.3 | - |

The pairwise comparison results are displayed in Tbl. 2, where each column presents the preference of one approach over the other methods. For example, users preferred our result 83.7% of the time over [13]. Notably almost 4 out of 5 users prefer our method over the prior methods, which well demonstrates that our method indeed produces better visual results. See Fig. 9 for some visual results.

It is also worth noting that the ranking of other methods differs in this test compared to the other two experiments. For example, Closed-Form Matting [22] is the lowest ranked method on alphamatting.com of the methods we compare here, yet to users it is preferable to all other methods except our own and [28]. On the other hand, while DCNN [8] is the prior state-of-the-art method on alphamatting.com, is only preferred over two methods on the real images. It is unclear whether this is due to methods overfitting the alphamatting.com dataset or whether the standard error metrics fail to accurately measure human perceptual judgment of alpha matting results.

## 6. Conclusion

In order to generalize to natural images, matting algorithms must move beyond using color as a primary cue and leverage more structural and semantic features. In this work, we show that a neural network is capable of capturing such high-order features and applying them to compute improved matting results. Our experiments show that our method does not only outperform prior methods on the standard dataset, but that it generalizes to real images significantly better as well.